\documentclass{article}

\usepackage{microtype}
\usepackage{graphicx}
\usepackage{subcaption}
\usepackage{booktabs} 

\usepackage{hyperref}

\usepackage[preprint]{icml2026}

\usepackage{amsmath}
\usepackage{amssymb}
\usepackage{mathtools}
\usepackage{amsthm}

\usepackage[capitalize,noabbrev]{cleveref}

\theoremstyle{plain}

\theoremstyle{definition}

\theoremstyle{remark}

\usepackage[textsize=tiny]{todonotes}

\icmltitlerunning{Chimera: Latency and Performance-Aware Multi-agent Serving for Heterogeneous LLMs}

\usepackage{enumitem}

\usepackage{booktabs}
\usepackage{tabularx}
\usepackage{multirow}
\usepackage{graphicx}
\usepackage{subcaption}
\usepackage{adjustbox}
\usepackage{xspace}

\makeatletter
\@ifpackageloaded{algorithmic}{%

}{}
\makeatother

\usepackage{algorithm}
\usepackage[noend]{algpseudocode}
\usepackage{amsmath}

\usepackage{dashrule}

\definecolor{commentblue}{RGB}{82,200,245}

\newcommand{\LineComment}[1]{\State \textcolor{commentblue}{// #1}}

\newcommand{\ours}{Chimera\xspace}

\begin{document}

\twocolumn[
  \icmltitle{\ours: Latency- and Performance-Aware Multi-agent Serving for Heterogeneous LLMs}




  \icmlsetsymbol{equal}{*}



    \begin{icmlauthorlist}
      \icmlauthor{Kangqi Ni}{unc}
      \icmlauthor{Wenyue Hua}{msr}
      \icmlauthor{Xiaoxiang Shi}{cmu}
      \icmlauthor{Jiang Guo}{amzn}
      \icmlauthor{Shiyu Chang}{ucsb}
      \icmlauthor{Tianlong Chen}{unc}
    \end{icmlauthorlist}
    
    \icmlaffiliation{unc}{University of North Carolina, Chapel Hill}
    \icmlaffiliation{msr}{Microsoft}
    \icmlaffiliation{cmu}{Carnegie Mellon University}
    \icmlaffiliation{amzn}{Amazon}
    \icmlaffiliation{ucsb}{University of California, Santa Barbara}

  \icmlcorrespondingauthor{Kangqi Ni}{kangqini@cs.unc.edu}
  \icmlcorrespondingauthor{Tianlong Chen}{tianlong@cs.unc.edu}

  \icmlkeywords{Machine Learning, ICML}

  \vskip 0.3in
]



\printAffiliationsAndNotice{}  

\begin{abstract}
Multi-agent applications often execute complex tasks as multi-stage workflows, where each stage is an LLM call whose output becomes part of context for subsequent steps. Existing LLM serving systems largely assume homogeneous clusters with identical model replicas. This design overlooks the potential of heterogeneous deployments, where models of different sizes and capabilities enable finer trade-offs between latency and performance. However, heterogeneity introduces new challenges in scheduling across models with diverse throughput and performance.
We present \ours, a predictive scheduling system for multi-agent workflow serving on heterogeneous LLM clusters that jointly improves end-to-end latency and task performance. \ours applies semantic routing to estimate per-model confidence scores for each request, predicts the total remaining output length of the workflow, and estimates per-model congestion using in-flight predicted token volumes for load balancing. 
%
We evaluate \ours on representative agentic workflows for code generation and math reasoning using multiple heterogeneous LLM configurations. Across comparable settings, \ours traces the best latency--performance frontier, reducing end-to-end latency by 1.2--2.4$\times$ and improving task performance by 8.0--9.5 percentage points on average over competitive baselines including vLLM.
\end{abstract}

\section{Introduction}

Large language models (LLMs) are increasingly deployed not only as standalone chatbots, but as multi-agent systems that solve tasks through collaborative workflows~\cite{plaat2025agentic_llms_survey, guo2024llm_multiagents_survey, wu2023autogen, hong2023metagpt, li2023camel}. In such systems, a single user query triggers a sequence of dependent LLM calls,
where each stage consumes the accumulated context and passes its output to the next stage. This shift from isolated requests to multi-agent workflows is widely adopted in diverse real-world applications~\cite{wu2023autogen, park2023generative_agents, qian2024chatdev, yang2024swe_agent}.

At the same time, LLM serving is both expensive and latency-sensitive, motivating extensive work on inference optimizations to improve throughput ~\cite{vllm23,sglang24,fu2024efficient,jiang2025cascadia}.
Meanwhile, deployments are becoming increasingly \emph{heterogeneous}: providers expose multiple model families and parameter sizes and use routing to adaptively select models per query, balancing cost and performance~\cite{ong2025routellm,feng2025ipr,shirkavand2025cscr,huang2025lookahead}.
However, most routing methods assume a simplified or static view of system load, treating latency as a fixed model attribute rather than an emergent outcome of queueing and resource contention under bursty traffic. Conversely, existing serving systems and schedulers primarily optimize per-request execution within a homogeneous model cluster (the same model replicated on each engine), and therefore do not address how heterogeneous model capacity should be allocated across requests of varying difficulty, or how model choice itself affects queue congestion.

\begin{figure}[t]
  \centering
  \includegraphics[
    width=\columnwidth,
    trim=0.1mm 0mm 2mm 1mm,
    clip
  ]{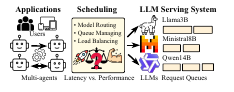}

    \vspace{-2mm}
    \caption{LLMs differ in latency and performance, and model choice directly affects queue congestion under load. Online serving over mixed LLMs requires a co-design of model routing, queue managing, and load balancing. How can we schedule multi-agent workflows to balance low latency and high performance?}
  \label{fig:motivation1}
  \vspace{-8mm}
\end{figure}

These assumptions are not suitable for serving multi-agent workflows on heterogeneous LLM clusters. First, workflow indicates end-to-end latency emerges from queueing interactions across multiple dependent stages, making request-level scheduling suboptimal~\cite{autellix25}. Second, model selection and congestion are tightly coupled: routing requests to slower but more capable models ensure performance but can amplify contention for shared GPU resources and degrade latency for other workflows. As a result, multi-agent serving exhibits an intrinsic trade-off between workflow end-to-end latency and task performance, which is not adequately addressed by prior work (Figure~\ref{fig:motivation1}).
Motivated by these observations, we ask:
\emph{How can we schedule multi-agent workflows over heterogeneous LLMs to improve the achievable latency and performance outcomes under load?}

We present \ours, a predictive scheduling system for multi-agent workflow serving on heterogeneous LLM clusters that jointly improves latency and performance.
\ours uses a transformer-based semantic router to estimate per-model confidence scores for each request, a lightweight CPU-based regressor to predict the total remaining output tokens of a workflow (current stage plus future stages), and an activity monitor that tracks in-flight predicted token volume per engine to estimate load-induced delay.
The load balancer then selects the best model that is sufficiently confident while remaining within a configurable latency slack of the fastest available option.
Requests are prioritized with a Shortest Total Job First (STJF) policy, with an aging-based anti-starvation mechanism.
\ours is implemented as a layer atop vLLM with asynchronous, batched router and predictor services to keep scheduling overhead low. Our contributions are as follows:
\begin{itemize}[leftmargin=*,itemsep=1pt, topsep=1pt, parsep=1pt, partopsep=1pt]

    \item \emph{Problem formulation for heterogeneous multi-agent serving.}
    We formulate online serving of multi-agent workflows on heterogeneous LLMs and measure the achievable \((\text{latency}, \text{performance})\) operating points under load.
        
  \item \emph{Predictive scheduling that couples model routing and workflow-level length prediction.}
  \ours combines semantic routing, workflow-level total output length prediction, and real-time load monitoring, using these signals to drive model selection and queue optimization.

  \item \emph{Latency and performance-aware dispatching.}
  We propose a dispatch rule that chooses the highest-confidence model within a latency slack of the fastest option, and prioritize workflows by predicted total length with aging to reduce latency while preventing starvation.

  \item \emph{End-to-end prototype and evaluation on agentic workflows.}
  We implement \ours atop vLLM and evaluate on code generation and math reasoning agentic tasks, showing lower latency and higher performance across diverse sets of LLMs comparing to baselines.
\end{itemize}




\section{Related Work}
\label{sec:related_work}

\textbf{LLM serving systems.}
vLLM~\cite{vllm23} improves KV-cache utilization via PagedAttention, and SGLang~\cite{sglang24} supports efficient execution of structured LM programs with RadixAttention for KV reuse and fast structured decoding.
Sarathi-serve optimizes batching with chunked prefill~\cite{sarathi_serve24}.
A broad line of work further improves throughput and latency via phase splitting and disaggregated prefill/decode~\cite{patel2023splitwise,distserve24,hu2024inference}. 
At the application layer, Parrot exposes cross-request structure for optimization~\cite{parrotserve24}, while Autellix targets agentic programs and reduces cumulative wait time with program-aware MLFQ scheduling~\cite{autellix25}.
These systems, however, assume homogeneous deployments, consisting of one or multiple replicas of the same model. Instead, we address heterogeneous multi-LLM serving by building a predictive scheduler on top of an existing serving engine (vLLM), co-designing routing, queue optimization, and load balancing to improve both latency and performance.

\textbf{Output length prediction for scheduling.}
Prior work has shown that predicting generation length can improve LLM scheduling and batching efficiency~\cite{jin2023s3,cheng2024lmaas,qiu2024proxy,stojkovic2024dynamollm}.
Most existing approaches formulate length prediction as request-level regression or classification~\cite{cheng2024lmaas,qiu2024proxy,qiu2024muserve,zheng2023perception}; recent work also demonstrates that relative length ordering alone can be sufficient for queue optimization~\cite{fu2024efficient}.
In this work, we adapt length prediction to multi-agent workflows by using a lightweight CPU-based regressor to predict the total output length of the workflow. This prediction serves as a priority signal for scheduling, combined with an aging-based anti-starvation mechanism, to reduce end-to-end latency.

\textbf{LLM routing.}
LLM routing selects among multiple LLMs to maintain task performance while reducing cost/latency and has rapidly moved from academic prototypes to widely deployed products~\cite{zhang2023modelspider,withmartian2025modelrouter,requestyai2025routing,microsoft2025azuremodelrouter}.
Lightweight routers focus on fast online decisions~\cite{hari2023tryage}, while representation- and structure-aware methods learn richer decision signals via contrastive objectives ~\cite{chen2024routerdc} or explicit relational structure ~\cite{feng2025graphrouter}.
Recent methods also study stronger training paradigms for routing policies and test-time compute allocation~\cite{zhang2025routerr1,ding2025bestroute,wang2025mixllm}, interpretable routing objectives~\cite{song2025irtrouter}, and techniques to generalize routing to newly introduced models at test time~\cite{jitkrittum2025universal,zhang2025capabilityinstructiontuning,zhuang2024embedllm}.
In our setting, routing is only one component in a broader online serving loop for multi-stage agent workloads.
We integrate routing with workflow-aware prioritization and real-time load balancing, so model choice and queue optimization are co-designed together.

\begin{figure*}[ht]
  \centering
  \includegraphics[width=0.85\textwidth]{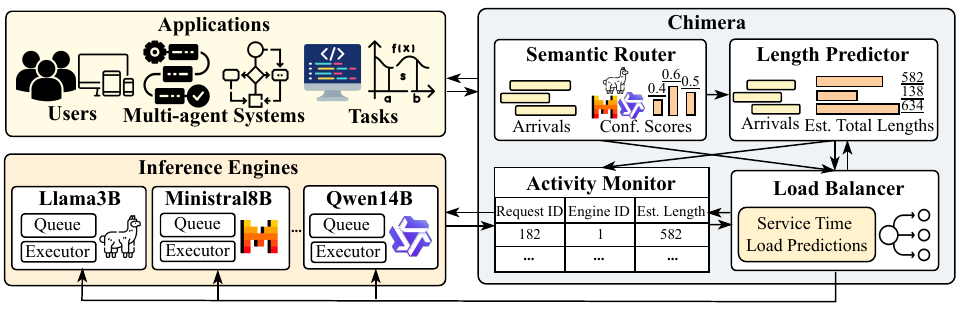}
  
    \caption{\ours system overview. 
    \ours is a middleware layer that sits between multi-agent applications and a pool of inference engines hosting heterogeneous LLMs of varying sizes and families.
    For each incoming request, \ours (i) predicts the confidence score for each candidate model (\textit{Semantic Router}), (ii) predicts the total number of output tokens (\textit{Length Predictor}), and (iii) tracks per-engine in-flight work (\textit{Activity Monitor}). It then combines these information to choose an engine and enqueue the request (\textit{Load Balancer}). Each backend engine executes requests from their local priority queues.}

  \label{fig:system}
\end{figure*}

\begin{algorithm}[t]
\caption{\ours's Predictive Scheduler}
\label{alg:predictive_scheduler}
\small
\begin{algorithmic}[1]

\Procedure{ScheduleRequest}{$r$, Models $\mathcal{M}$, AssignmentMap $\mathcal{A}$, InFlightTokens $\mathcal{I}$, ModelQueues $\mathcal{Q}$, ModelProfiles $\Pi$, Router $S$, Predictor $\hat{Y}$, LatencySlack $\tau$, ConfidenceMargin $\Delta s$}

\If{$r.\mathrm{program\_id} \notin \mathcal{A}$}

  \LineComment{Predict the confidence score for each model}
  \For{$m \in \mathcal{M}$}
    \State $q[m] \gets S(r.\mathrm{prompt}, m)$
  \EndFor

  \LineComment{Accumulate the TTLT for each model}
  \For{$m \in \mathcal{M}$}
    \State $P_m \gets 0$
    \For{\textbf{each} $t \in \mathrm{Values}(\mathcal{I}[m])$}
      \State $P_m \gets P_m + t$
    \EndFor
    \State $L[m] \gets \dfrac{P_m \cdot \Pi[m].\mathrm{decode\_ms\_per\_token}}{\Pi[m].\mathrm{max\_batch\_size}}$
  \EndFor
  \State $m_{\mathrm{fast}} \gets \arg\min_{m \in \mathcal{M}} L[m]$
  \State $L_{\mathrm{fast}} \gets L[m_{\mathrm{fast}}]$

  \LineComment{Select the highest scoring model within the latency slack and above the score margin}
  \State $m^\star \gets m_{\mathrm{fast}}$
  \State $q^\star \gets q[m_{\mathrm{fast}}]$
  \For{$m \in \mathcal{M}$ \textbf{sorted by} $q[m]$ \textbf{descending}}
    \If{$L[m] \le (1+\tau)\cdot L_{\mathrm{fast}}$}
      \If{$q[m] \ge q^\star + \Delta s$}
        \State $m^\star \gets m$
        \State $q^\star \gets q[m]$
        \State \textbf{break}
      \EndIf
    \EndIf
  \EndFor

  \State $\mathcal{A}[r.\mathrm{program\_id}] \gets m^\star$

\Else
  \LineComment{Reuse the assigned model}
  \State $m^\star \gets \mathcal{A}[r.\mathrm{program\_id}]$
\EndIf

\LineComment{Predict the total output tokens as the priority}
\State $\hat{y} \gets \hat{Y}(r.\mathrm{meta\_info}, m^\star)$
\State $r.\mathrm{predicted\_tokens} \gets \hat{y}$
\State $r.\mathrm{priority} \gets \hat{y}$

\LineComment{Track in-flight tokens and enqueue the request}
\State $\mathcal{I}[m^\star][r.\mathrm{id}] \gets r.\mathrm{predicted\_tokens}$
\State $\mathrm{Dispatch}(\mathcal{Q}[m^\star], r, r.\mathrm{priority}, r.\mathrm{arrival})$
\EndProcedure

\end{algorithmic}
\end{algorithm}

\section{Methodology}

\subsection{Problem Settings}
\textbf{Setups.} We consider a GPU cluster that serves multi-agent workflows. A workflow specification defines a sequence of agent stages, where each stage has a role (e.g., planner, solver, coder, verifier) and a system prompt encoding stage-specific instructions. The cluster deploys a heterogeneous set of LLMs $\mathcal{M}$ that differ in latency, memory footprint, and task performance; each call invocation is served as an LLM request routed to one of these models.

\textbf{Notations.} Each workflow execution is identified by a program id $p$ and consists of $N_p$ ordered stages $\{1,\dots,N_p\}$. The input to stage $i$ includes the stage-$i$ system prompt, the user message, and the accumulated history from prior stages. Let $C_p$ denote the end-to-end completion time of workflow $p$ (i.e., the completion time of stage $N_p$), and let $S_p$ denote a task-specific performance score of the final output (e.g., accuracy, pass rate). A request is a tuple
$\; r = (p, i, x_r, \textit{meta\_info}_r), \;$
where $x_r$ is the stage input prompt and $\textit{meta\_info}_r$ contains workflow identifiers and decoding parameters.

\textbf{Objective.} Our goal is to jointly improve the latency and performance of multi-agent workflows on a heterogeneous model cluster. Choosing a stronger model can increase $S_p$ but may increase $C_p$, especially under load. 
Because applications differ in how they value latency relative to performance, we avoid collapsing the two into a single scalar objective. 
Instead, we assess a serving system by the set of achievable operating points in terms of $(C, S)$: a system is strictly better if it can attain a lower latency with a higher performance metric at the same time.

\subsection{Overview}
\ours is implemented as a middleware layer between agent applications and backend inference engines (Figure~\ref{fig:system}). Applications define multi-agent workflows and issue requests that \ours schedules on the available inference engines.
The system is organized into four modules:
\begin{enumerate}[leftmargin=*,itemsep=1pt, topsep=1pt, parsep=1pt, partopsep=1pt]
  \item \textit{Semantic Router}: a transformer-based encoder that outputs a confidence score for each LLM request, estimating which model is most likely to achieve high performance.
  \item \textit{Length Predictor}: a regression model that estimates the total output token length for the workflow, which is used to determine the request’s scheduling priority.
  \item \textit{Activity Monitor}: a program table that tracks per-request metadata and maintains estimates of in-flight workload, measured in predicted token volume, for each engine.
  \item \textit{Load Balancer}: a dispatcher that integrates router confidence scores with the monitored per-engine service workload to select a target engine and dispatch the request accordingly.
\end{enumerate}

\textbf{Scheduling Algorithm.}
Upon receiving a request, \ours first determines the model that will execute it. If the request belongs to an ongoing workflow whose earlier stages have already been assigned to a model, that assignment is reused to preserve execution locality. Otherwise, \ours proceeds in two steps: it computes per-model scores using the semantic router, producing a logit for each candidate model that reflects predicted task performance, and it estimates each model’s current load based on the predicted token volume of requests already in flight. Model selection balances latency and performance. \ours chooses the highest-scoring model whose estimated delay falls within a configurable slack relative to the fastest available option, while also enforcing a minimum score margin to avoid switching models for marginal gains and to promote KV-cache reuse. After selecting a model, \ours predicts the total output tokens for the request’s workflow and uses this estimate as its scheduling priority, so that requests with smaller predicted total work are served earlier. Finally, the predicted token volume is recorded as additional in-flight load for the chosen model, and the request is inserted into that model’s priority queue for execution by the backend engine.

\begin{figure}[t]
  \centering
  \includegraphics[width=\columnwidth]{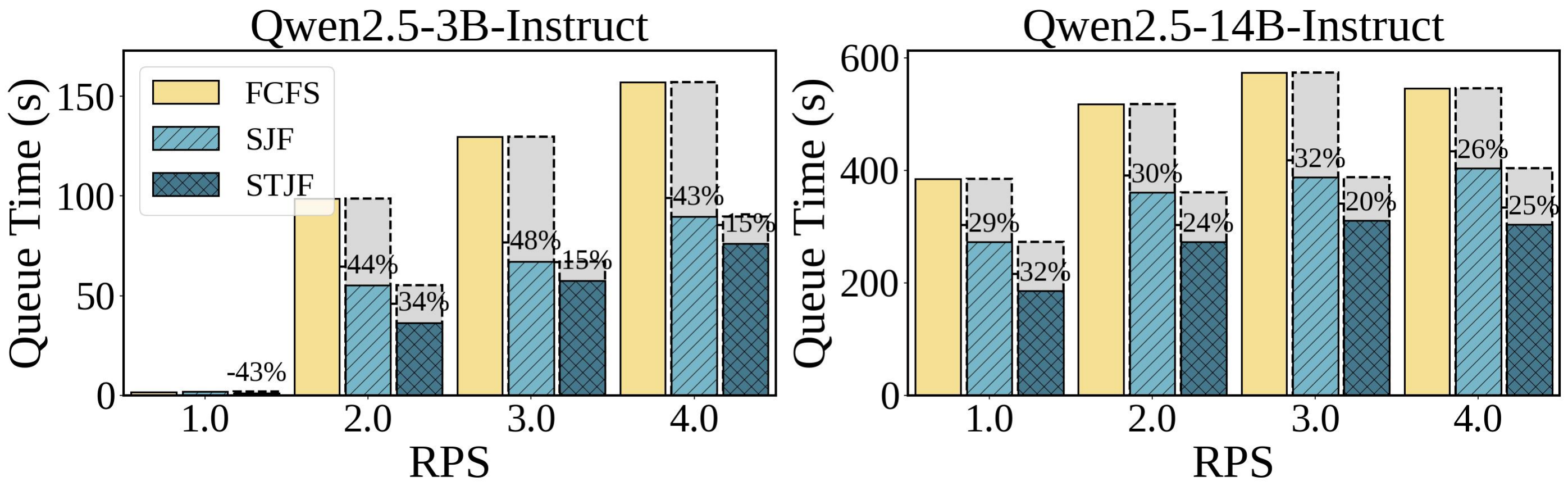}
  \caption{Average queue time for serving multi-agentic workflows with single Qwen models. Prioritization based on Shortest Total Job First (STJF) reduces the most amount of queue time, comparing to First-Come First-Served (FCFS) and Shortest Job First (SJF).}
  \label{fig:motivation2}
  \vspace{-6mm}
\end{figure}

\subsection{Semantic Router}
\label{subsec:router}
\textbf{Objective.} For each incoming request $r$, the router outputs a vector of confidence scores $\{q[m]\}_{m\in\mathcal{M}}$, where $q[m]\in[0,1]$ estimates the probability that model $m$ successfully solves $r$. The load balancer uses these confidence scores to select a target model.

\textbf{Modeling.} The semantic router $S$ is a fine-tuned transformer encoder from ModernBert-large \cite{warner2025smarter} with a multi-label head that produces independent confidence scores per candidate model. Given a tokenized prompt $x_r$, the encoder produces a pooled representation $h_r$ (e.g., [CLS]), and a linear head outputs logits $\ell_{r,m}$ for each model. We use sigmoid activations to obtain $q[m]=\sigma(\ell_{r,m})$. Training uses offline traces where each $(x_r,m)$ pair is labeled by task-specific correctness (1 if correct, 0 otherwise), optimized with binary cross entropy.

\subsection{Length Predictor}
\label{subsec:predictor}
\textbf{Objective.} For a request $r=(p,i,\cdot)$ in a multi-agent workflow, we characterize the total work as the sum of outputs produced by the current and subsequent agent stages.
We define the total output tokens as
$\, Y_r \;=\; \sum_{j=i}^{N_p} \text{OutTokens}(p,j), \,$
where $\text{OutTokens}(p,j)$ is the output length of stage $j$.
\ours predicts this quantity online and attaches the estimate $\hat{y}\approx \text{median}(Y_r \mid \phi(r), m^\star)$ to each request, where $m^\star$ is the selected model.
The predicted $\hat{y}$ serves as the scheduling priority: requests with smaller estimated total work are given higher priority, approximating a Shortest Total Job First (STJF) policy.
While one could prioritize by a weighted combination of input (prefill) tokens and predicted output (decode) tokens, we find that it provides no measurable improvement over using $\hat{y}$ alone as decoding time significantly dominates prefill time for long outputs.
As illustrated in Figure~\ref{fig:motivation2}, Shortest Job First (SJF), which schedules requests without workflow awareness, reduces queueing time by 26--48\% relative to First-Come First-Served (FCFS).
Incorporating workflow-level information through STJF yields an additional 15--34\% reduction compared to SJF.

\textbf{Modeling.}
\ours uses a Quantile Random Forest (QRF) to predict the conditional median of $Y_r$. QRF offers two practical advantages: (i) low resource consumption and cheap CPU-based inference, and (ii) robustness to heavy-tailed token lengths via quantiles.
The predictor's features are designed to be informative about both agent workflow and prompt complexity and inexpensive to be computed online. We use the following features:
\begin{itemize}[leftmargin=*,itemsep=1pt, topsep=1pt, parsep=1pt, partopsep=1pt]
  \item \textit{Token counts:} token counts of the system prompt, user prompt, and full input prompt.
  \item \textit{Workflow state:} categorical identifiers for the multi-agent workflow and the current agent stage.
  \item \textit{Model name:} the base model identity since output length distributions vary by model.
  \item \textit{Text sketch:} a compact representation of user prompt via TF-IDF (uni/bi-grams) followed by dimensionality reduction through truncated Singular Value Decomposition.
\end{itemize}

\subsection{Activity Monitor}
\textbf{Objective.} The activity monitor maintains a program table that tracks per-request metadata and execution state, including workflow identifiers, the assigned engine identifier, and the estimated total remaining output length. It also maintains the predicted in-flight token volume used for subsequent load estimation.

For each served model $m$, the monitor stores a map $\mathcal{I}[m]$ for each request id and its predicted token volumes. Each request in the system has one entry $r.\mathrm{id}\mapsto \hat{y}$, where $\hat{y}$ is produced by the length predictor after model selection. The entry is removed once the request completes. Aggregating these values provides an estimate of per-engine service-time workload, and these aggregates are exposed to the load balancer to inform routing decisions.

\subsection{Load Balancer}
\textbf{Objective.} The load balancer selects an engine that improves expected performance while controlling load-induced delay, and enqueues the request with a priority that reflects predicted total amount of work. It receives (i) router scores $\{q[m]\}_{m\in\mathcal{M}}$ and (ii) activity-monitor state $(\mathcal{A},\mathcal{I})$, and dispatches the request to a selected model $m^\star$ with the predicted length as the priority.

\textbf{Load estimation from in-flight predicted tokens.}
When no prior assignment exists, the load balancer estimates the time-to-last-token for each model using the accumulated in-flight predicted tokens from previously admitted requests. For model $m$, it sums the token volumes in $\mathcal{I}[m]$:
$\; P_m = \sum_{t\in \mathrm{Values}(\mathcal{I}[m])} t, \;$
and computes the corresponding TTLT estimate
$\; L[m] = \dfrac{P_m \cdot \Pi[m].\mathrm{decode\_ms\_per\_token}}{\Pi[m].\mathrm{max\_batch\_size}} \;$ where $decode\_ms\_per\_token$ is profiled offline.
This estimate accounts for both queue length and expected token volume already admitted to each model.

\textbf{Latency and performance-aware routing.}
The load balancer identifies the fastest model $m_{\mathrm{fast}}=\arg\min_{m\in\mathcal{M}} L[m]$ with $L_{\mathrm{fast}}=L[m_{\mathrm{fast}}]$. 
It then selects the highest-confidence model within a configurable latency slack: starting from $m^\star=m_{\mathrm{fast}}$ and $q^\star=q[m_{\mathrm{fast}}]$, it iterates models in descending $q[m]$ and updates $m^\star$ to the first model satisfying both (i) a load constraint $L[m]\le (1+\tau)\cdot L_{\mathrm{fast}}$ and (ii) a confidence improvement $q[m]\ge q^\star+\Delta s$. 
This rule yields monotone fallback behavior under load: as a larger model becomes congested, its $L[m]$ increases and the system shifts traffic to less-loaded models. If the request's program id has an existing assignment, we reuse that assignment to reduce the overhead of calling the semantic router for multiple times, which occupies GPU resources; at the same time, this application of the router does not require a complex training process, thus reduces the complexity of developing the router.

\textbf{Priority assignment.}
After selecting the model $m^\star$, \ours invokes the length predictor to obtain $\hat{y}=\hat{Y}(r.\mathrm{meta\_info}, m^\star)$
and uses $\pi_r:=\hat{y}$ as the request priority. It then (i) inserts $r.\mathrm{id}\mapsto \hat{y}$ into the in-flight map $\mathcal{I}[m^\star]$ and (ii) pushes the request into the per-model queue $\mathcal{Q}[m^\star]$ keyed by $(\pi_r, r.\mathrm{arrival})$, 
where arrival order breaks ties (FCFS among equal predicted work). The requests preserve their priority scores within each inference engine's priority queue.

\subsection{Anti-starvation.} 
Pure priority scheduling can delay long requests when short requests continue to arrive. \ours mitigates this with an aging-based promotion mechanism controlled by two parameters: a starvation threshold $S$ and a running quantum $Q$. This is implemented within each inference engine. Each queued request maintains an $starvation\_count$ recording how many consecutive scheduling iterations it was not selected; once $starvation\_count \ge S$, the request is temporarily promoted by setting its $starvation\_level - 1$, causing it to sort ahead of all normal requests in the min heap based priority queue.
Promoted requests remain in this elevated starvation level. After receiving $Q$ scheduling iterations, the requests are then demoted down a starvation level ($starvation\_level + 1$), restoring standard ordering. 
The priority queue key is $(starvation\_level, priority, arrival\_time)$ within a local engine. This simple mechanism improves worst-case waiting time while preserving the intended priority rule,

\subsection{Implementation Details}
\ours builds on top of vLLM and exposes the same API compatible with standard OpenAI chat and generation endpoints. Furthermore, to keep per-request overhead low, both the semantic router and the length predictor run as asynchronous batched services: incoming requests enqueue lightweight feature objects, a background collector forms batches up to a maximum size or a timeout; batched inference is executed in a separate worker pool so that process-based isolation avoids contending with the scheduler event loop. The scheduler only blocks on the minimal outputs needed for dispatch (router confidence scores for model selection, and an estimated total output length for priority), then records the prediction into the in-flight accounting and enqueues the request to the chosen engine. Inference engines remain responsible for continuous batching, kernel execution, and KV-cache management.

\begin{figure}[t]
  \centering
  \begin{subfigure}[t]{0.5\columnwidth}
    \centering
    \includegraphics[width=\linewidth]{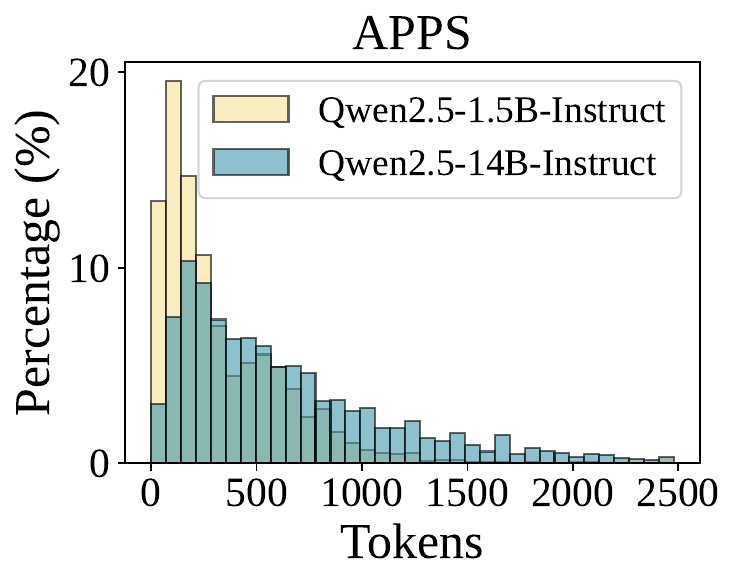}

    \label{fig:outlen-apps}
  \end{subfigure}\hfill
  \begin{subfigure}[t]{0.5\columnwidth}
    \centering
    \includegraphics[width=\linewidth]{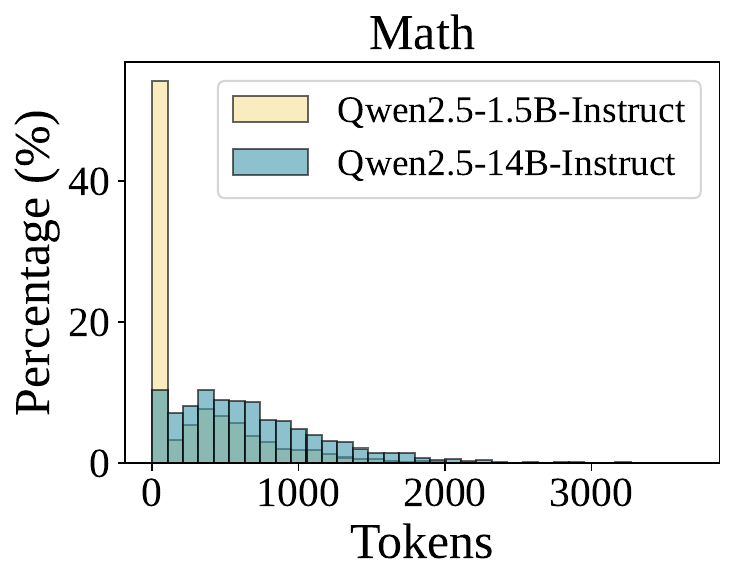}

    \label{fig:outlen-math}
  \end{subfigure}

  \caption{Total output length distributions for APPS and MATH. We observe that different models exhibit different total output length distributions with large standard deviations.}
  \label{fig:traces}

  \vspace{-7mm}

\end{figure}

\begin{figure*}[!htbp]
  \centering

  \includegraphics[
    width=\textwidth,
    trim=0 0.95cm 0 0,
    clip
  ]{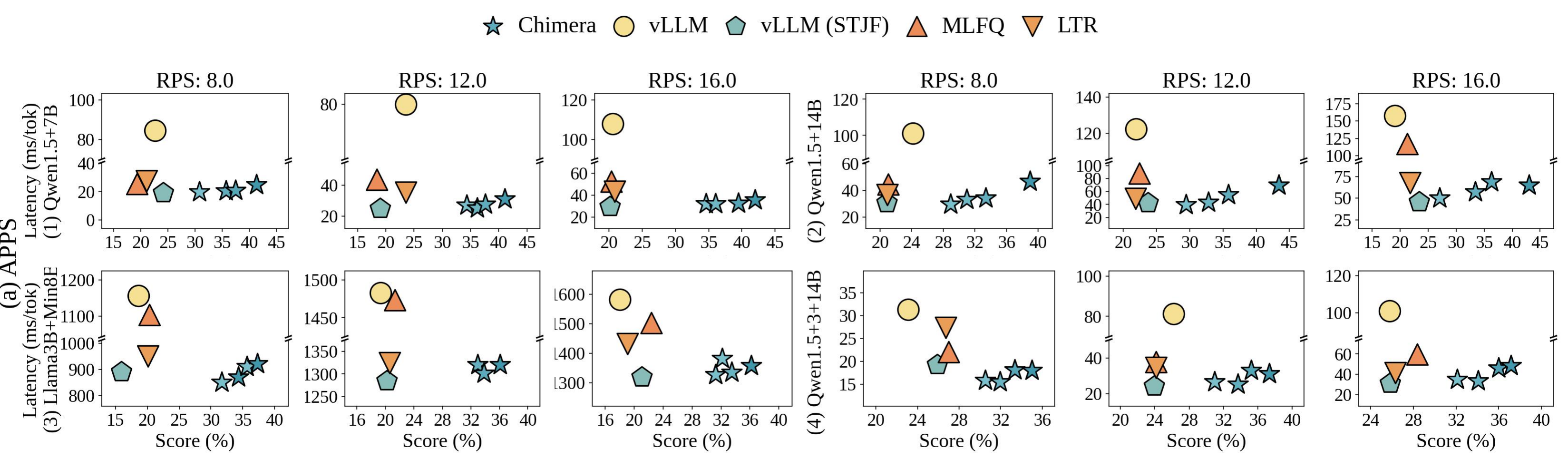}

    \par\vspace{-1.5mm}
    \noindent\hdashrule{\textwidth}{0.3pt}{1.2pt 1.2pt}\par
    \vspace{2.0mm}

  \includegraphics[
    width=\textwidth,
    trim=0 1mm 0 3.2cm,
    clip
  ]{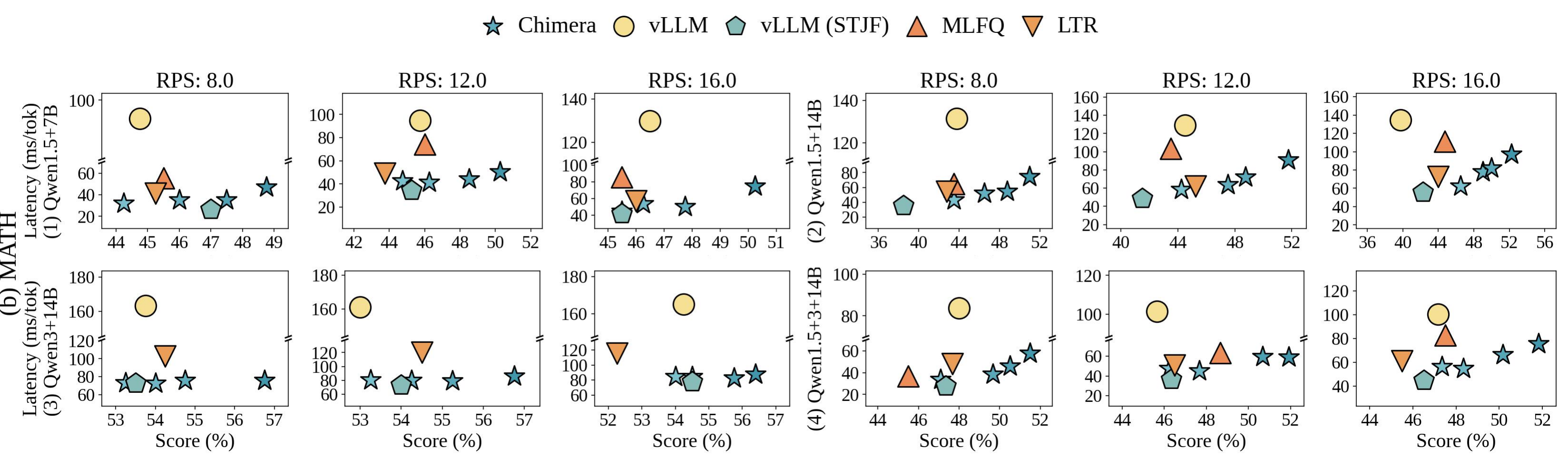}

  \caption{Latency vs. performance on heterogeneous model combinations. We evaluate on LLMs with various parameter sizes (1.5B, 3B, 7B, 8B, and 14B) and families (Qwen, Llama, and Ministral). 
  \ours demonstrates lower latency and higher performance compared to other serving systems.}

  
  \label{fig:main_results}

  \vspace{-3.5mm}
\end{figure*}

\section{Experiments}

\subsection{Datasets}
\label{subsec:datasets}

We evaluate \ours with representative agentic tasks, such as code generation and mathematical reasoning, which vary in the total output token lengths across models as shown in Figure~\ref{fig:traces}. We construct workflows per dataset based on ReAct~\citep{yao2022react}; and each task's workflow requires between one to four stages to complete. At runtime, the user chooses a workflow to execute a query. 
This simulates a realistic multi-agent serving situation in practice.

\textbf{Code Generation: APPS.} The APPS dataset is a benchmark of programming problems drawn from competitive-programming and interview-style settings. Each example provides a natural-language problem statement with input/output format details and constraints and requires generating an executable Python solution that passes hidden test cases.
For Qwen2.5-1.5B-Instruct, the total output length averages $447$ tokens with a standard deviation of $1276$ tokens. For Qwen2.5-14B-Instruct, the total output length averages $649$ with a standard deviation of $534$ tokens.

\textbf{Math Reasoning: MATH.} The MATH dataset is a collection of competition-style mathematics problems covering diverse topics (e.g., algebra, geometry, number theory, combinatorics, probability) and multiple difficulty levels. Problems are typically presented in short-form natural language and require producing a final numeric or symbolic answer. The dataset includes questions from common contest sources, such as AMC 10, AMC 12, and AIME.
For Qwen2.5-1.5B-Instruct, the total output length averages $606$ tokens with a standard deviation of $2587$ tokens. For Qwen2.5-14B-Instruct, the total output length averages $709.0,$ with a standard deviation of $715$ tokens.

\subsection{Experimental Setup}
We evaluate \ours across popular LLM families and scales: Qwen2.5-1.5B/3B/7B/14B-Instruct (Qwen1.5B/3B/7B/14B)~\cite{qwen2025qwen25technicalreport}, Ministral3-8B-Instruct (Minis8B)~\cite{mistral_ministral3_8b_instruct_2512_bf16}, and Llama3.2-3B-Instruct (Llama3B)~\cite{grattafiori2024llama}.
We use these shortened model names in later sections for brevity.
For each dataset, we evaluate four deployment configurations that vary in model family, scale, and the number of deployed models. All experiments run on NVIDIA RTX A6000 GPUs (49GB VRAM).

\textbf{Baselines.}
We evaluate \ours against three state-of-the-art LLM serving systems:
\begin{itemize}[leftmargin=*,itemsep=1pt, topsep=1pt, parsep=1pt, partopsep=1pt]
    \item vLLM: A high-performing LLM serving backend with PagedAttention and chunked prefill. Its load balancer dispatches each request to the engine with the lowest weighted sum of waiting and running requests.

    \item MLFQ: Extends vLLM with multiple priority queues and applies promotion and demotion based on service time.

    \item LTR: Uses a decoder-only language model to predict the relative ordering of request output lengths, and applies a SJF policy to prioritize requests within each vLLM engine’s local queue.
\end{itemize}
We also include using the oracle workflow-level total output length for prioritization in vLLM as a reference--vLLM(STJF)--to observe the lowest possible latency.

\textbf{Metrics.}
We focus on measuring both the end-to-end (E2E) latency of workflows and the task performance. Therefore, we report the average end-to-end latency per token vs. the task-specific performance score. For code generation, the score is the average \% tests passed; for mathematical reasoning, the score is the average exact match. 
Points closer to the lower-right corner of the plot are better, as they show lower latency with higher performance.

\subsection{Main Results}
\newcommand{\pct}[1]{#1\%}                 
\newcommand{\speedup}[1]{#1$\times$}       
\newcommand{\mspt}[1]{#1\,ms/token}        
\newcommand{\rps}[1]{RPS=#1}               

In Figure~\ref{fig:main_results}, we present the main evaluation results of \ours. Across settings, \ours achieves the strongest operating points—simultaneously attaining the lowest E2E latency and the highest performance. 
Since other baselines do not jointly optimize E2E latency and performance for heterogeneous LLM serving, their gains primarily manifest as limited reductions in end-to-end latency, with little or no improvement in task performance.
Moreover, \ours provides a tuning knob via the latency slack threshold, enabling providers to systematically adjust the latency--performance trade-off: tighter slack prioritizes lower latency, while larger slack unlocks higher performance.

\textbf{APPS.}
On APPS (Figure~\ref{fig:main_results}(a1)--(a4)), \ours consistently achieves the strongest operating points across model configurations. 
For Qwen1.5+7B (Figure~\ref{fig:main_results}(a1)), across three RPS values (\rps{8,12,16}), vLLM achieves per token E2E latencies of \mspt{82.6}, \mspt{81.0}, and \mspt{107.8}, with corresponding performance scores of \pct{22.7}, \pct{23.5}, and \pct{20.6}. MLFQ achieves an average \speedup{2.4} speedup but incurs a \pct{2.9} average drop in scores over vLLM. LTR is more effective, achieving an average \speedup{2.6} speedup with the same score over vLLM. In comparison, \ours improves both latency and performance, achieving a \speedup{3.4} speedup and a \pct{16.0} score increase over vLLM, a \speedup{1.42} speedup and a \pct{19.6} score increase over MLFQ, and a \speedup{1.31} speedup and a \pct{16.0} score increase over LTR. Notably, increasing the latency slack improves performance without degrading latency, illustrating a minimal latency--performance tradeoff. For Qwen1.5+14B (Figure~\ref{fig:main_results}(a2)), MLFQ achieves an average \speedup{1.7} speedup over vLLM, while LTR achieves an average \speedup{2.5} speedup over vLLM. \ours achieves the strongest joint improvement, attaining a \speedup{2.9} speedup and a \pct{10.5} score increase over vLLM, a \speedup{1.7} speedup and a \pct{10.8} score increase over MLFQ, and a \speedup{1.2} speedup with a \pct{10.9} score increase over LTR. For Llama3B + Ministral8B (Figure~\ref{fig:main_results}(a3)), the overall latency is substantially higher because Llama3B and Ministral8B produces significantly longer outputs than the other models (Appendix~\ref{app:traces_more}). vLLM incurs per-token latencies of \mspt{1156.7}, \mspt{1482.7}, and \mspt{1582.6} for \rps{8,12,16} respectively. MLFQ provides little latency reduction in this setting. LTR reduces latency under lighter loads (\rps{8,12}) but still struggles at a higher load (\rps{16}). In contrast, \ours achieves the strongest overall outcome, with a \speedup{1.2} speedup and a \pct{15.7} score increase over vLLM, a \speedup{1.2} speedup and a \pct{13.0} score increase over MLFQ, and a \speedup{1.1} speedup and a \pct{14.4} score increase over LTR. Since vLLM (STJF) only achieves a \speedup{1.2} speedup over vLLM, this shows that \ours already achieves near oracle speedup. We further evaluate \ours on the three-model configuration Qwen1.5+3+14B (Figure~\ref{fig:main_results}(a4,b4)). Compared to two-model combinations, introducing a third model adds additional routing and load-balancing dynamics due to a wider range of latency--performance profiles. On APPS, MLFQ achieves low latency under light load (\rps{8.0}) but struggles as load increases (\rps{12,16}). LTR achieves better speedup while maintaining the same score over vLLM. \ours achieves a \speedup{2.7} speedup with a \pct{8.2} score increase over vLLM, a \speedup{1.5} speedup with a \pct{6.7} score increase over MLFQ, and a \speedup{1.3} speedup with a \pct{7.5} score increase over LTR.

\textbf{MATH.}
On MATH, \ours similarly achieves the strongest latency--performance trade-offs across the evaluated model configurations. 
For Qwen1.5+7B (Figure~\ref{fig:main_results}(b1)), LTR is second to \ours, achieving an average \speedup{2.2} speedup and the same score over vLLM. In contrast, \ours achieves a \speedup{2.5} speedup and a \pct{2.2} score increase over vLLM, as well as a \speedup{1.2} speedup and a \pct{2.1} score increase over LTR. With a larger latency slack, \ours offers a trade-off for better performance, achieving an average \speedup{1.9} speedup with a \pct{4.1} score increase over vLLM. For Qwen1.5+14B (Figure~\ref{fig:main_results}(b2)), the improvement also persists on MATH: \ours achieves an average \speedup{2.1} speedup with a \pct{5.0} score increase over vLLM, a \speedup{1.4} speedup and a \pct{3.8} score increase over MLFQ, the same latency and a \pct{3.7} score increase over LTR. Increasing latency slack of \ours can further improve the performance with a small tradeoff of latency. For Qwen1.5+3+14B (Figure~\ref{fig:main_results}(b4)), \ours achieves a \speedup{2.1} speedup and a \pct{1.6} score increase over vLLM, a \speedup{1.2} speedup and a \pct{1.3} score increase over MLFQ, and a \speedup{1.2} speedup and a \pct{1.2} score increase over LTR.

\textbf{Scheduling overhead.}
Table~\ref{tab:scheduling_overhead} breaks down \ours's scheduling overhead across datasets, RPS, and model combinations. Overall, \ours adds negligible overhead: the scheduler contributes at most \(2.2\%\) of end-to-end latency (and often \(<1\%\)). On APPS it ranges from \(0.7\%\)–\(2.2\%\) and on MATH from \(0.2\%\)–\(0.4\%\), showing that \ours is extremely lightweight.

\begin{table}[t]
\centering
\scriptsize
\setlength{\tabcolsep}{3.8pt}
\renewcommand{\arraystretch}{1.10}
\caption{\ours's scheduling overhead. The scheduler (containing the router, the predictor, and other operations altogether) contribute \(\le 2.2\%\) of the E2E latency across settings. This shows that \ours is lightweight.}

\resizebox{\columnwidth}{!}{%
\begin{tabular}{ll l r l l l}

\toprule
\textbf{Dataset} & \textbf{RPS} & \textbf{Models} & \textbf{E2E Latency} &
\textbf{Router} & \textbf{Predictor} & \textbf{Scheduler} \\
\midrule

\multirow{6}{*}{\textbf{APPS}}
  & \multirow{3}{*}{8.0}
    & Qwen1.5+7B      & 17068 &  64 (0.4\%) & 179 (1.1\%) &  250 (1.5\%) \\
  & & Qwen1.5+14B     & 28750 &  72 (0.3\%) & 126 (0.4\%) &  206 (0.7\%) \\
\cmidrule(lr){2-7}
  & \multirow{3}{*}{16.0}
    & Qwen1.5+7B      & 24554 & 277 (1.1\%) & 234 (1.0\%) &  528 (2.2\%) \\
  & & Qwen1.5+14B     & 46522 & 315 (0.7\%) & 315 (0.7\%) &  646 (1.4\%) \\
\midrule

\multirow{6}{*}{\textbf{MATH}}
  & \multirow{3}{*}{8.0}
    & Qwen1.5+7B      & 37025 &  49 (0.1\%) &  90 (0.2\%) & 144 (0.4\%) \\
  & & Qwen1.5+14B     & 61944 &  44 (0.1\%) &  82 (0.1\%) & 131 (0.2\%) \\
\cmidrule(lr){2-7}
  & \multirow{3}{*}{16.0}
    & Qwen1.5+7B      & 45485 &  71 (0.2\%) &  78 (0.2\%) & 155 (0.3\%) \\
  & & Qwen1.5+14B     & 79485 &  52 (0.1\%) & 101 (0.1\%) & 159 (0.2\%) \\
\bottomrule
\end{tabular}%
}

\label{tab:scheduling_overhead}

\vspace{-3mm}
\end{table}

\subsection{Ablations}
We conduct ablations to analyze the contributions of \ours's key decision modules in isolation: the semantic router and the length predictor.

\begin{figure}[t]
  \centering
  \includegraphics[width=\columnwidth, trim=0.8cm 0cm 0 0, clip]{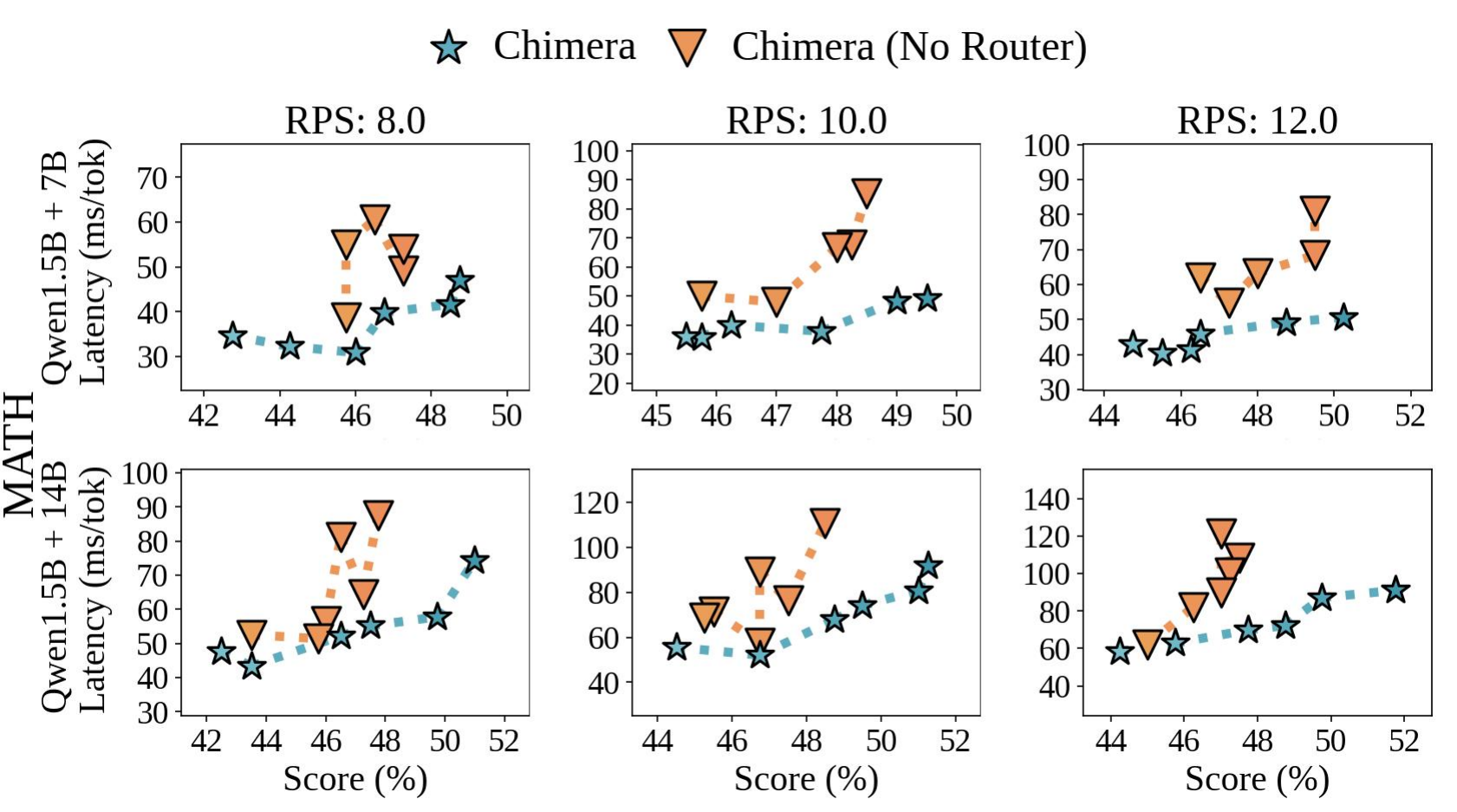}

    \caption{\small Pareto frontiers of \ours with and without the semantic router. We ablate the router and compare the resulting latency--performance tradeoff curves. Using the router yields a markedly better Pareto frontier across all RPS values.}

    \vspace{-2mm}
  \label{fig:ablation_router}
\end{figure}

\textbf{Effect of the semantic router.} 
We compare \ours with and without the router in Figure~\ref{fig:ablation_router}. We find that the router confidence scores enable \ours to choose which model to run for each query based on the expected performance benefit, rather than applying a fixed heuristic (e.g., always using the stronger model when the latency slack permits). This query-adaptive selection yields a markedly better Pareto frontier.
Without a semantic router, the system lacks per-query confidence scores and therefore incurs higher latency to achieve comparable performance gains. This supports our design principle of coupling per-query, confidence-based model routing with predictive scheduling to jointly improve latency and performance.

\begin{figure}[t]
  \centering
  \includegraphics[width=\columnwidth, trim=0.8cm 0 0 0, clip]{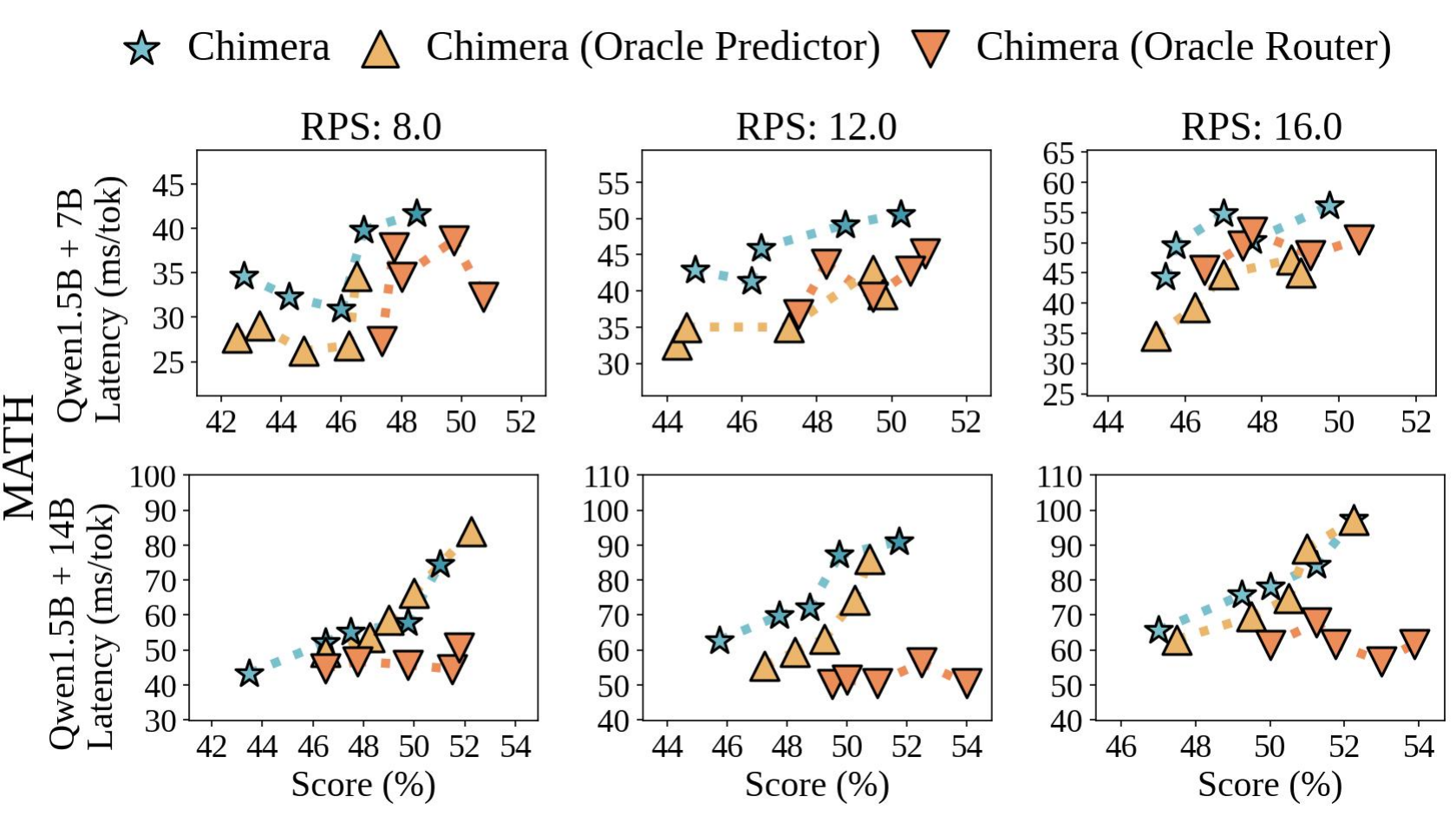}


\caption{\small Pareto frontiers of \ours and its oracle variants: (i) Oracle Predictor uses ground-truth total output lengths, and (ii) Oracle Router uses ground-truth per-request success outcomes. The benefit of each oracle dependends on the model combination.}

  \label{fig:oracle}

  \vspace{-6mm}
\end{figure}

\textbf{Comparison with the oracle.}
To quantify the remaining headroom in our learned components, we construct oracle variants of \ours by replacing (i) the length predictor with ground-truth generation lengths (Oracle Predictor) and (ii) the router with ground-truth per-request model success outcomes (Oracle Router) in Figure~\ref{fig:oracle}. 
An oracle predictor further reduces latency for Qwen1.5+7B especially given low latency slacks, indicating that length-prediction error matters most when latency constraints are tight; however, it has limited impact for Qwen1.5+14B, where \ours performs nearly identically with and without the oracle at RPS=8 and 16. 
An oracle router improves the latency--performance frontier for Qwen1.5+14B, yielding high performance at low latency, but offers smaller gains for Qwen1.5+7B, suggesting routing is not the primary bottleneck there. 
These oracle studies provide upper bounds: \ours is already near-oracle in several settings, and the remaining gaps highlight the main opportunities for improvement (length prediction for Qwen1.5+7B and routing for Qwen1.5+14B).


\section{Conclusion}

We present \ours, a lightweight middleware for serving multi-agent workflows on heterogeneous LLM clusters that jointly optimizes routing, load-aware dispatch, and workflow-aware prioritization to achieve strong latency--performance trade-offs. \ours combines a ModernBERT-based semantic router for per-request model confidence, a CPU-efficient QRF predictor for workflow-level remaining output tokens (enabling STJF scheduling), and an activity monitor that estimates load via in-flight predicted token volume, with an aging-based anti-starvation mechanism to bound waiting time. Across APPS and MATH under varying RPS values and diverse model combinations, \ours consistently improves end-to-end latency while also increasing task performance, achieving favorable Pareto frontiers relative to vLLM, MLFQ, and LTR, while adding negligible scheduling overhead.

\section*{Acknowledgement}
This project was funded in part by the Amazon Research Award. The views and conclusions contained in this document are those of the authors and should not be interpreted as representing official policies, either expressed or implied, of the funding organizations.

\bibliography{references}
\bibliographystyle{icml2026}

\newpage
\appendix
\onecolumn







\section{Trace Files}

\subsection{Agentic Workflows}
\label{app:workflows}

We use common agentic workflows based on the widely adopted ReAct framework. For code generation, we use three types of workflows that require between one to four stages: (1) Planner $\rightarrow$ Coder $\rightarrow$ QAAggent $\rightarrow$ Coder (4 stages); (2) Planner $\rightarrow$ Coder (2 stages); (3) Coder (1 stage). For mathematical reasoning, we also use three types of workflows from one to four stages: (1) Planner $\rightarrow$ Solver $\rightarrow$ Verifier $\rightarrow$ Solver (4 stages); (2) Planner $\rightarrow$ Solver (2 stages); (3) Solver (1 stage). These workflows represent a varying number of stages, which is common in multi-agent applications.

\subsection{Total Output Length}
\label{app:traces_more}
We show the total workflow-level output lengths for each model on the datasets to compliment the discussions in Section~\ref{subsec:datasets}. This motivates our design for having model-specific workflow-level total output length predictors. Since we use CPU-based QRFs, it is cheap to train and serve multiple predictors simultaneously.

\begin{figure}[H]
  \centering
  \begin{subfigure}[t]{0.24\columnwidth}
    \centering
    \includegraphics[width=\linewidth]{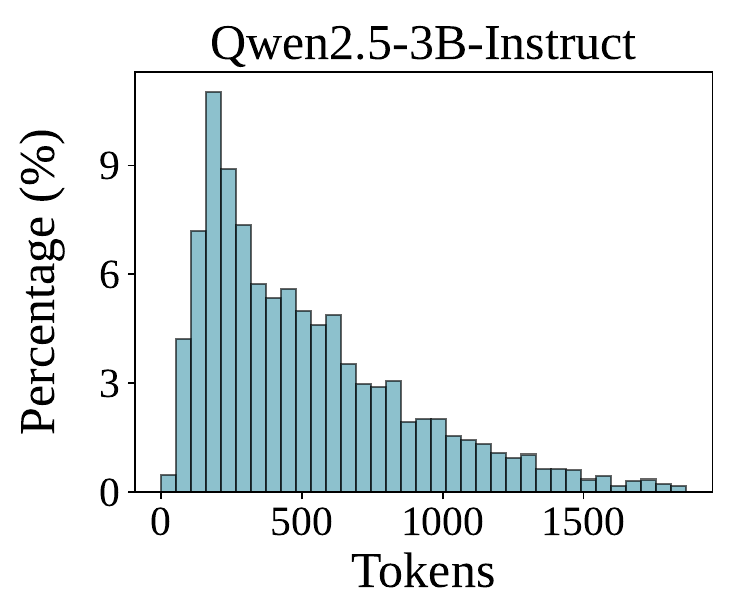}
    \label{fig:}
  \end{subfigure}\hfill
  \begin{subfigure}[t]{0.24\columnwidth}
    \centering
    \includegraphics[width=\linewidth]{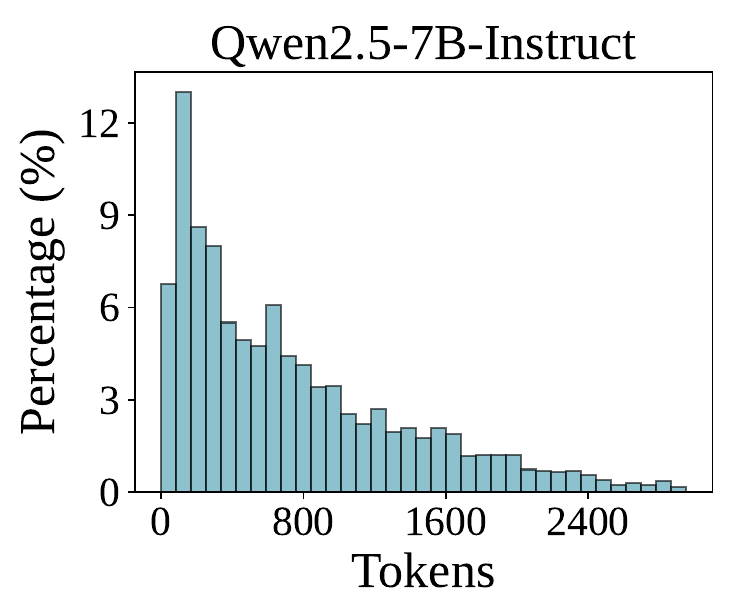}
    \label{fig:outlen-math-1}
  \end{subfigure}\hfill
  \begin{subfigure}[t]{0.24\columnwidth}
    \centering
    \includegraphics[width=\linewidth]{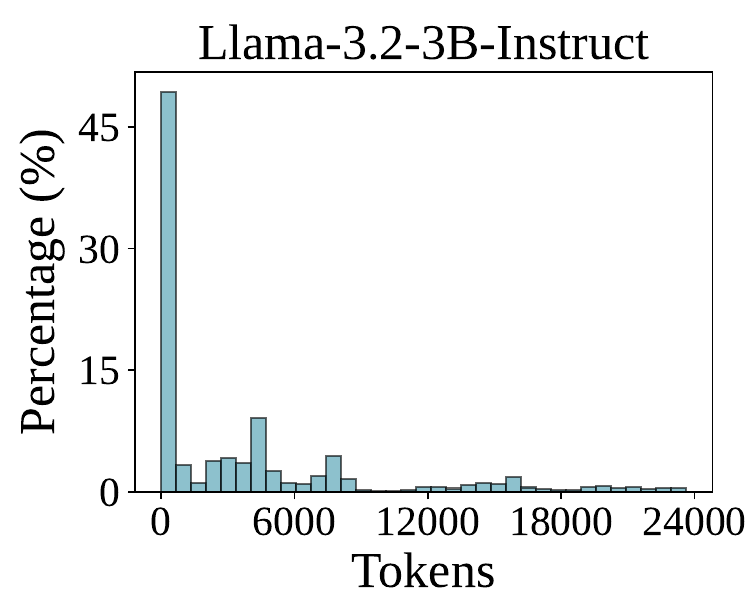}
    \label{fig:outlen-apps-2}
  \end{subfigure}\hfill
  \begin{subfigure}[t]{0.24\columnwidth}
    \centering
    \includegraphics[width=\linewidth]{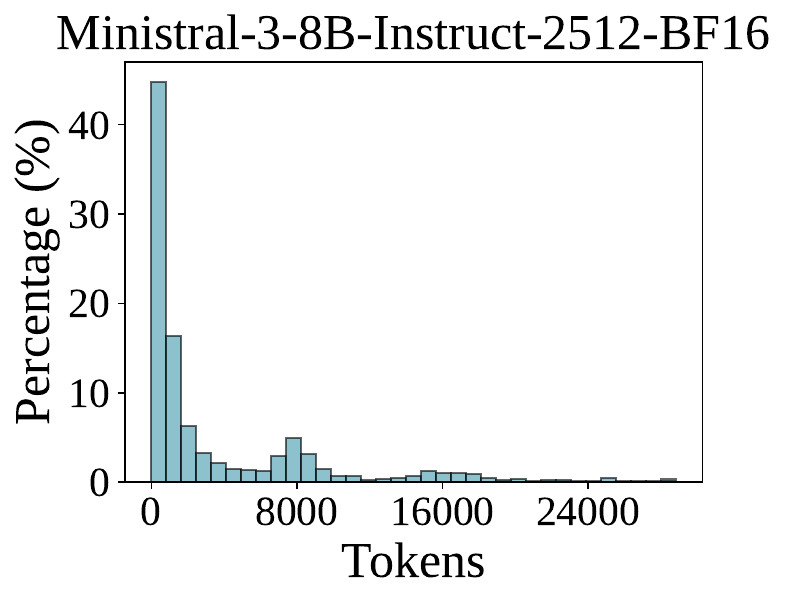}
    \label{fig:outlen-math-2}
  \end{subfigure}

  \vspace{-5mm}
  
  \caption{Total output length distributions for APPS.}
  \label{fig:traces}
  \vspace{-4mm}
\end{figure}

\begin{figure}[H]
  \centering
  \begin{subfigure}[t]{0.25\columnwidth}
    \centering
    \includegraphics[width=\linewidth]{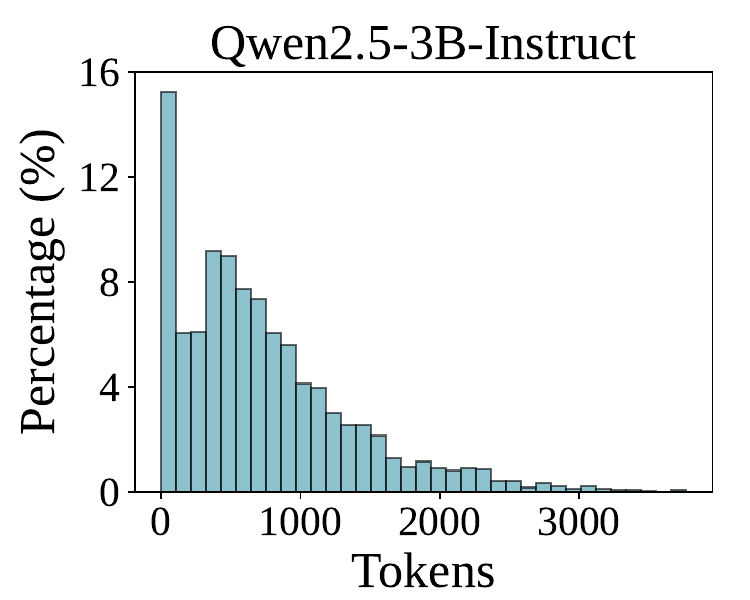}
    \label{fig:}
  \end{subfigure}\hfill
  \begin{subfigure}[t]{0.25\columnwidth}
    \centering
    \includegraphics[width=\linewidth]{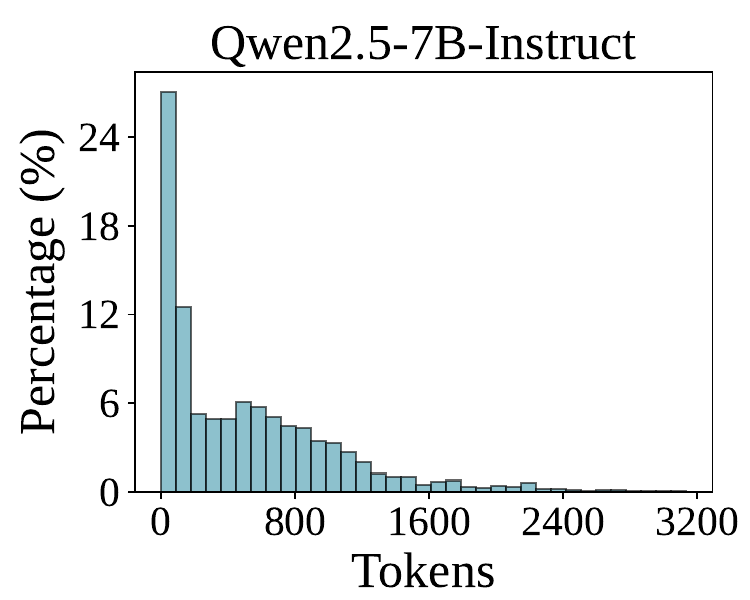}
    \label{fig:outlen-math-1}
  \end{subfigure}\hfill
  \begin{subfigure}[t]{0.25\columnwidth}
    \centering
    \includegraphics[width=\linewidth]{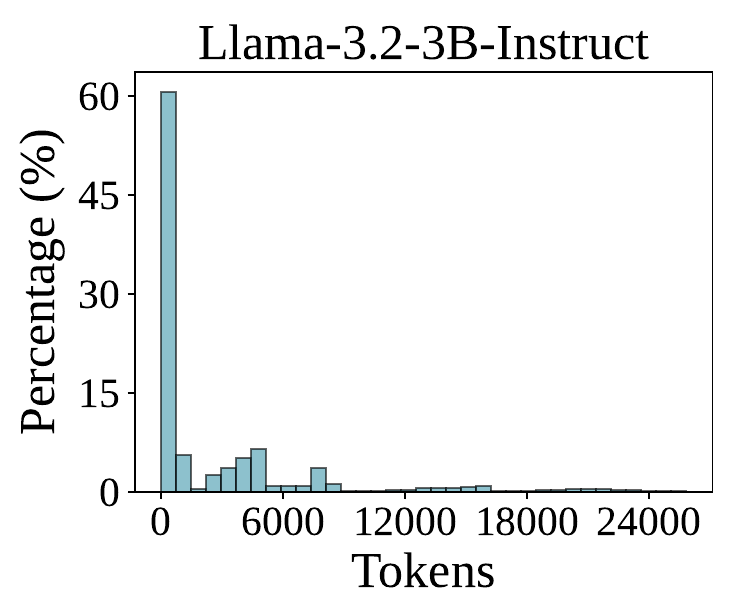}
    \label{fig:outlen-apps-2}
  \end{subfigure}\hfill
  \begin{subfigure}[t]{0.25\columnwidth}
    \centering
    \includegraphics[width=\linewidth]{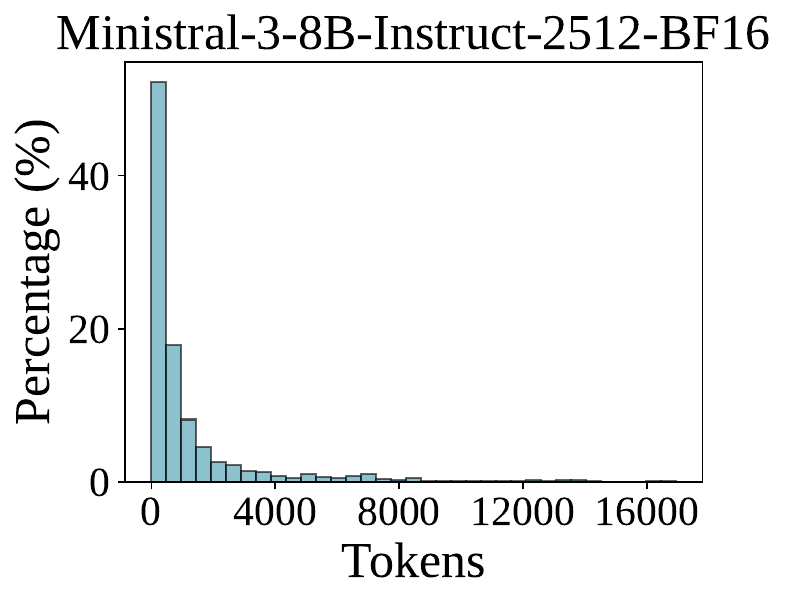}
    \label{fig:outlen-math-2}
  \end{subfigure}

   \vspace{-5mm}

  \caption{Total output length distributions for MATH.}
  \label{fig:more_traces}
  \vspace{-4mm}
\end{figure}

\section{Module Analysis}

\subsection{Semantic Router}
We finetune the semantic router for $5$ epochs, a learning rate of $1e-5$, and a batch size of $16$. The training objectives have been detailed in Section~\ref{subsec:router}.
Since we directly leverage the router confidence scores in making decisions, we evaluate the router's performance using macro mean Average Precision (meanAP) as threshold-free metric. From the results, we select ModernBert-large as the backbone as supports longer context than the original Bert models and achieves the best meanAP.

\begin{table}[H]
\centering
\small
\renewcommand{\arraystretch}{1.15}

\caption{Performance of the semantic router.}
\adjustbox{max width=\columnwidth}{%
\begin{tabular}{l l c}
\toprule
Predictor & Dataset & mAP \\
\midrule
\multirow{3}{*}{ModernBert-large}
  & APPS & 0.521 \\
  & MATH & 0.740 \\
  & Avg. & 0.631 \\
\midrule
\multirow{3}{*}{ModernBert-base}
  & APPS & 0.474 \\
  & MATH & 0.716 \\
  & Avg. & 0.595 \\
\bottomrule
\end{tabular}
}

\label{tab:predictor_eval}
\end{table}

\subsection{Length Predictor}
The training objectives have been detailed in
Section~\ref{subsec:predictor}.
We measure the predictor's performance as relative ranking errors using Kendall's Tau distance $\in [0,1]$ (lower is better) comparing with the oracle STJF ranking in Table~\ref{tab:predictor_eval}. For instance, a 0.124 means around 12.4\% of the pairs are inverted in the ranking, while the oracle is 0\% and FCFS is around 50\% inversions. 
We show that using the predictor can markedly reduce Kendall’s tau distance by 0.314 over FCFS and 0.371 over using the input length as a proxy.

\begin{table}[t]
\centering
\small
\renewcommand{\arraystretch}{1.15}

\caption{Performance of total output length predictors.}

\adjustbox{max width=\columnwidth}{%
\begin{tabular}{l l c c c c c c}
\toprule
Predictor & Dataset 
& Qwen1.5B 
& Qwen3B 
& Qwen7B 
& Qwen14B 
& Llama3B 
& Ministral8B \\
\midrule
\multirow{3}{*}{QRF}
  & APPS & 0.207 & 0.139 & 0.124 & 0.129 & 0.270 & 0.220 \\
  & MATH & 0.146 & 0.195 & 0.185 & 0.222 & 0.254 & 0.154 \\
  & Avg.  & 0.176 & 0.167 & 0.155 & 0.175 & 0.262 & 0.187 \\
\midrule
\multirow{3}{*}{FCFS}
  & APPS & 0.501 & 0.506 & 0.506 & 0.505 & 0.505 & 0.496 \\
  & MATH & 0.498 & 0.495 & 0.495 & 0.498 & 0.506 & 0.495 \\
  & Avg.  & 0.499 & 0.501 & 0.500 & 0.502 & 0.505 & 0.496 \\
\midrule
\multirow{3}{*}{Input Length}
  & APPS & 0.567 & 0.468 & 0.442 & 0.442 & 0.489 & 0.489 \\
  & MATH & 0.704 & 0.646 & 0.670 & 0.642 & 0.525 & 0.609 \\
  & Avg.  & 0.635 & 0.557 & 0.556 & 0.542 & 0.507 & 0.549 \\
\bottomrule
\end{tabular}
}

\label{tab:predictor_eval}
\end{table}

\section{Extra Main Results}
\begin{figure*}[ht]
  \centering

  \includegraphics[
    width=\textwidth,
    trim=0 1cm 0 0,
    clip
  ]{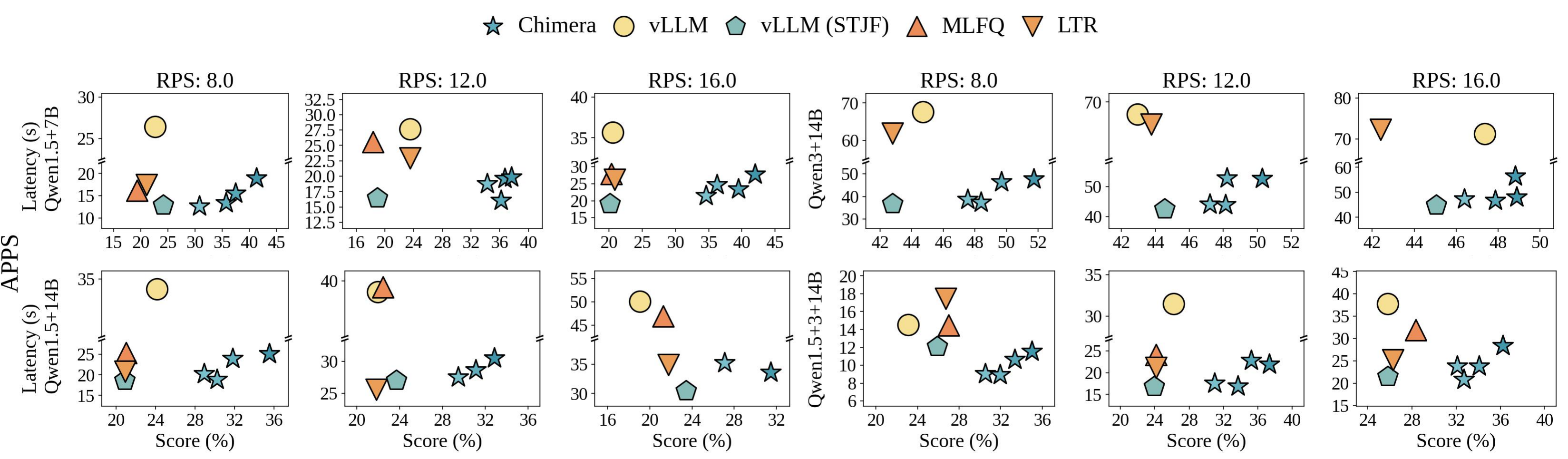}

  \vspace{2mm}

  \includegraphics[
    width=\textwidth,
    trim=0 1mm 0 3.2cm,
    clip
  ]{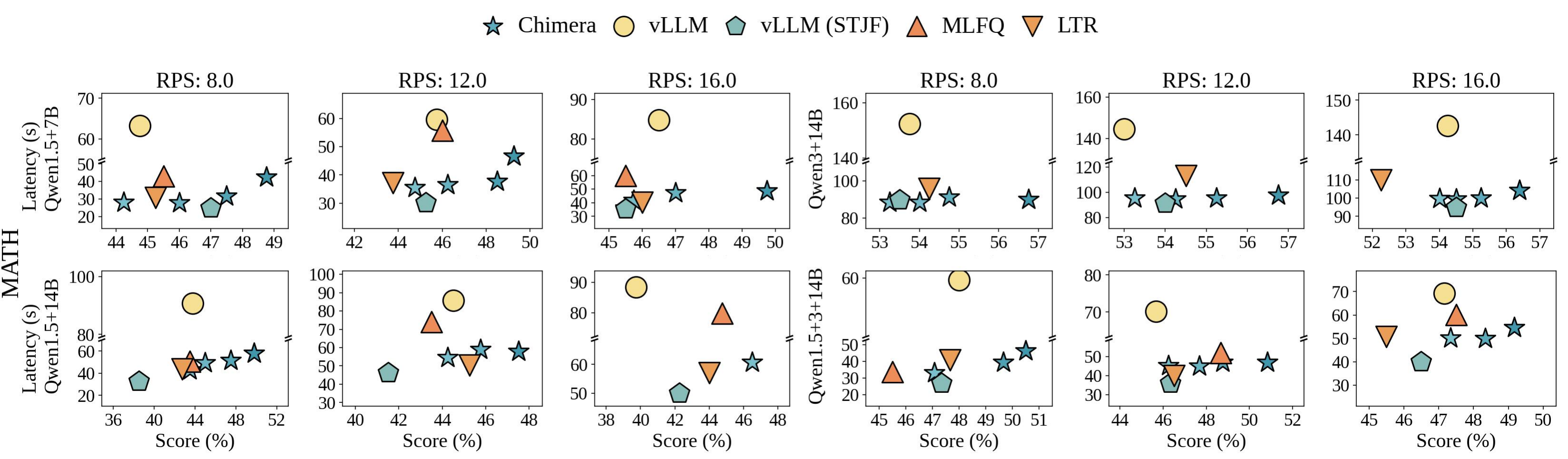}

  \caption{Latency vs. performance on heterogeneous model serving combinations.}
  
  \label{fig:main_results_e2e}
\end{figure*}

Besides end-to-end latency per token, we present extra main results with the average end-to-end latency per workflow (makespan) in Figure~\ref{fig:main_results_e2e}. \ours achieves the lowest makespan and the highest performance across all settings.

\end{document}